# Equirectangular image construction method for standard CNNs for Semantic Segmentation

Haoqian Chen[1], Jian Liu[2], Minghe Li[1], Kaiwen Jiang[1], Ziheng Xu[1], Rencheng Sun[1] and Yi Sui [1,*]

[1] College of Computer Science and Technology, University of Qingdao, Qingdao, 266071, China
[2] Qingdao Stomatological Hospital, Qingdao, 266000, China
* Correspondence: suiyi@qdu.edu.cn

**Abstract:** 360° spherical images have advantages of wide view field, and are typically projected on a planar plane for processing, which is known as equirectangular image. The object shape in equirectangular images can be distorted and lack translation invariance. In addition, there are few publicly dataset of equirectangular images with labels, which presents a challenge for standard CNNs models to process equirectangular images effectively. To tackle this problem, we propose a methodology for converting a perspective image into equirectangular image. The inverse transformation of the spherical center projection and the equidistant cylindrical projection are employed. This enables the standard CNNs to learn the distortion features at different positions in the equirectangular image and thereby gain the ability to semantically the equirectangular image. The parameter, φ, which determines the projection position of the perspective image, has been analyzed using various datasets and models, such as UNet, UNet++, SegNet, PSPNet, and DeepLab v3+. The experiments demonstrate that an optimal value of φ for effective semantic segmentation of equirectangular images is 6π/16 for standard CNNs. Compared with the other three types of methods (supervised learning, unsupervised learning and data augmentation), the method proposed in this paper has the best average IoU value of 43.76%. This value is 23.85%, 10.7% and 17.23% higher than those of other three methods, respectively.

## 1. Introduction

360° spherical images have increasingly wide applications in the fields of automated driving, drones, VR, and more[1]. Equidistant cylindrical projection is often used to map a 360° spherical image to a 2D plane. As shown in Fig.1, the distorted shape of the object can be clearly observed in the equirectangular image. The closer to the upper and lower ends of the image, the more serious the distortion of the shape of the object. However, the distortion of the shape of the object in the middle area is lighter[2,3]. Different locations in the equirectangular image show different object shapes and structural features, which pose challenges for object detection, semantic segmentation, and depth estimation of equirectangular images.

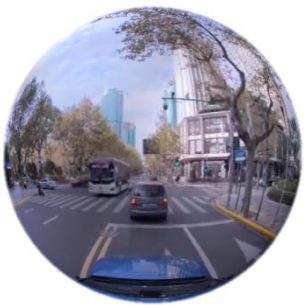
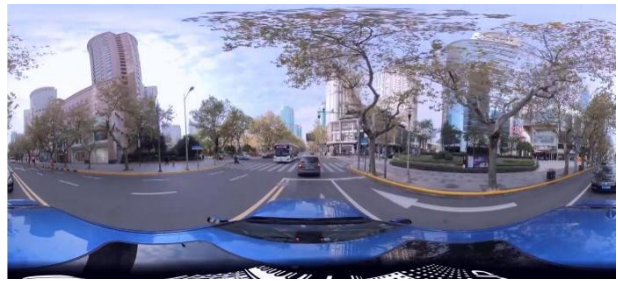

**Figure 1.** Equirectangular image obtained by equidistant projection of 360° image

At present, Convolutional Neural Networks (CNNs) have excellent performance in image classification, image segmentation, target recognition, and other fields. Especially based on large-scale labeled datasets (COCO[4], ImageNet[5], Cityscapes[6]), pre-trained CNN models such as VGG[7] and ResNet[8] can be used as initial models for solving new problems and tuned on new datasets. The data objects processed by CNNs are usually perspective images, which are characterized by the same object having the same shape in different locations (i.e., translational invariance). Thus, the convolution kernel of CNNs uses regular squares. However, equirectangular images do not have translational invariance characteristics. To solve this problem, some studies have proposed the method of distorting the convolution kernel by adaptively adjusting its shape according to the degree of distortion of the object shape. Coors et al.[9] and Tateno et al.[10] changed the shape of the convolution kernel through spherical projection. As shown in Fig. 2, the closer to the middle of the equirectangular image, the closer the shape of the convolution kernel is to a square; the closer to the top and bottom of the equirectangular image, the left and right boundaries of the convolution kernel are inclined to the sides to adapt to changes in the shape of objects in this area. The use of distorted convolutional kernels is premised on supervised learning, which assumes the existence of a large labeled dataset of equirectangular images.

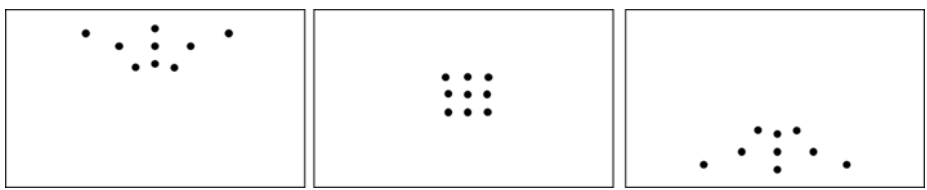

**Figure 2.** Adaptive adjustment of convolutional kernel shape of CNNs

After the widespread adoption of 360° cameras, there has been a significant increase in the volume of 360° spherical images and videos available. Despite this, there remains a shortage of labeled equirectangular image data, primarily due to the costly nature of annotation. To address this issue, several studies have implemented domain adaptive methods aimed at processing equirectangular images. In their study, Su et al.[11] utilized the tangent plane of the perspective image and equirectangular image to train a convolutional neural network (CNN). They used a pre-trained CNN for the tangent plane of the perspective image, while for the equirectangular image, they employed a CNN with parameters that needed to be optimized. They continuously optimized the parameters of the latter so that the output would be similar to that of the pre-trained CNN. Although this method does not require a labeled equirectangular images dataset, the model has a large number of parameters (up to GB level), making it difficult to train and use. The method described by the authors[12] has been optimized to reduce the number of model parameters. However, the effectiveness of the improved method is determined by the performance of the pre-trained model. If the pre-trained model fails to perform well, it can negatively affect the effectiveness of the model on the equirectangular image. Ma et al.[13] proposed a semantic segmentation model for equirectangular images using generative adversarial networks (GANs), which is similar to the domain adaptive method. The model takes perspective images with labels and equirectangular images as inputs, which then undergo semantic segmentation through a segmentation network (Generator G) to produce segmentation results. The objective is to minimize the disparity between the two segmentation results. To achieve this, the segmentation results are evaluated using the discriminator network (D) to classify them as either perspective images or equirectangular images. Generator G is trained using segmentation loss and backpropagation loss to generate segmentation results comparable to those of perspective images. The method proposed in this study does not rely on labeled equirectangular images. However, it specifically focuses on the central region of the equirectangular image, which experiences less distortion. The study does not investigate the upper and lower regions of the image that are known to exhibit significant distortion. Our findings indicate that the method is not effective in handling the severely distorted portions of equirectangular images.

Training a network to process equirectangular images using a limited perspective image dataset can be challenging, especially when extracting shape features from severely distorted objects in these images. To overcome this challenge, we present a novel approach to transform perspective image datasets into labeled equirectangular image datasets. This conversion method enables standard CNN models to effectively process equirectangular images, including the upper and lower regions where object shape features are highly distorted. We propose an optimal projection position for converting perspective image datasets into equirectangular image datasets for semantic segmentation. Our proposed method surpasses other existing approaches, demonstrating superior performance.

## 2. Method

### 2.1. Coordinate transformation of spherical images and equirectangular images

As depicted in Fig 3(a), the viewpoint of the spherical image is at the center of the sphere, denoted as $S$, which is a unit sphere. A point $s(\theta,\varphi)$ on the sphere can be uniquely determined by the azimuthal angle $\theta$ with the range of $[-\pi,\pi]$ and the zenith angle $\varphi$ within the range of $[-\frac{\pi}{2},\frac{\pi}{2}]$. Equidistant cylindrical projection samples the spherical image at equal intervals ($\Delta_\theta$ and $\Delta_\varphi$). These sampling points are evenly mapped onto a planar image, as shown in Fig.3(b). Each point on the sphere S corresponds to a point on the plane.

Suppose the width and the height of the plane are denoted as $w$ and $h$, respectively. Then, $\Delta_\theta = \frac{2\pi}{w}$, and $\Delta_\varphi = \frac{\pi}{h}$ represent the horizontal and vertical lengths of each unit on the plane. When processing planar images, it is common to convert the spherical coordinate system to a Cartesian coordinate system. As illustrated in Figure 3(c), a point $s(\theta,\varphi)$ on the sphere corresponds to a point $p(x,y)$ on the planar image, and the coordinate conversion are equations are as follows:

$$x = \frac{(\theta+\pi)w}{2\pi} \quad (1)$$

$$y = \frac{(\pi-2\varphi)h}{2\pi} \quad (2)$$

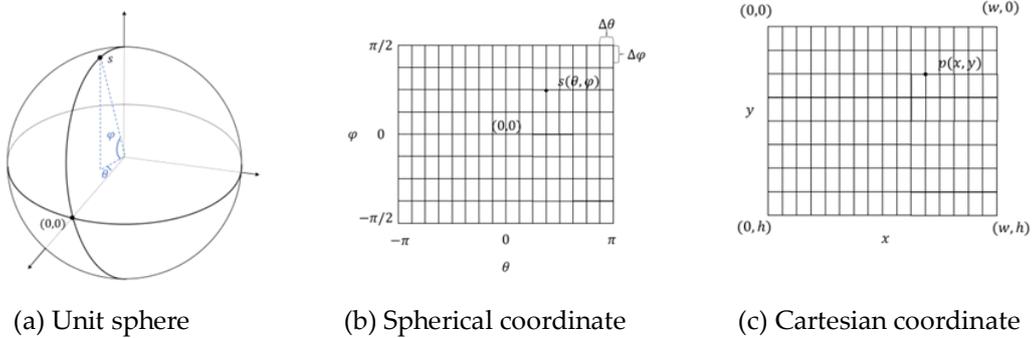

(a) Unit sphere      (b) Spherical coordinate      (c) Cartesian coordinate

**Figure 3.** Spherical image coordinate system and equidistant cylindrical projection

### 2.2. Spherical images and tangent planes

For any point on a sphere, it is possible to draw a tangent line that intersects the sphere. The point of intersection between the tangent plane and the sphere is known as the tangent point. In Fig.4(a), we select the point (0,0) on the spherical surface as the tangent point and construct a tangent plane, denoted as $I_r$, with size of $n \times n$ at (0, 0). By utillizing the spherical center projection[14], we can obtain a projection region, denoted as $I_s$, on the sphere with size of $n \times n$. Fig.4(b) illustrates the one-to-one correspondence between the points on $I_r$ and the points on $I_s$. This means that the orange tangent plane is projected to the blue area on the sphere.

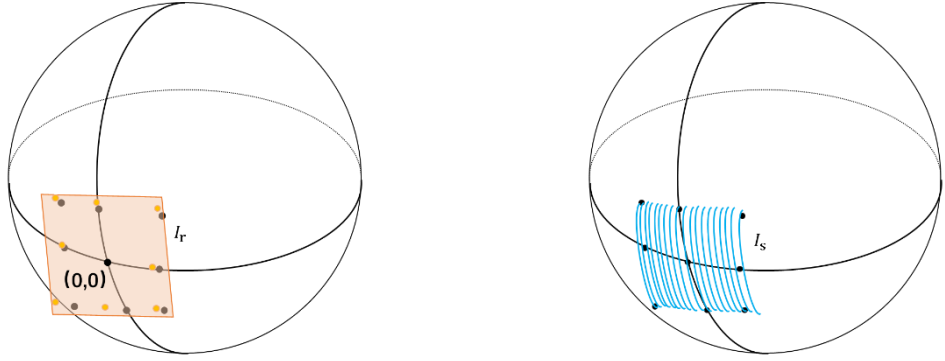

(a) Tangent plane $I_r$  (b) The projection region $I_s$

**Figure 4.** Tangent plane $I_r$ and its projection $I_s$ region on the sphere

where $i, j \in \left[0, \frac{n-1}{2}\right], n \geq 3$. The tangent $I_r$ at sampling intervals of $\Delta_\theta$ and $\Delta_\varphi$, in both horizontal and vertical directions is depicted in Fig.5. To determine the Cartesian coordinates of a point $r(i,j)$ on the tangent plane $I_r$, Equ.3~Equ.6 can be utilized. Here, $i$ and $j$ are within the range $\left[0, \frac{n-1}{2}\right]$, with $n$ being greater than or equal to 3.

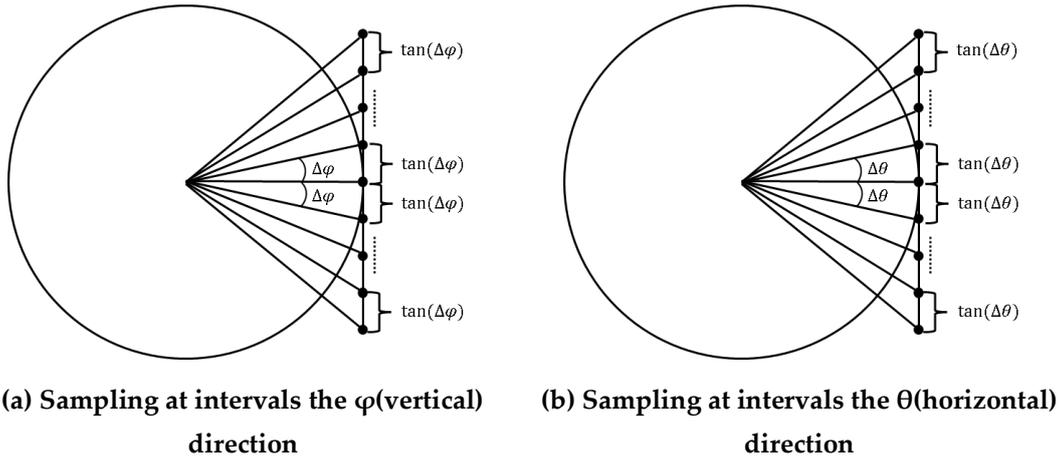

**(a) Sampling at intervals the $\varphi$(vertical) direction**  **(b) Sampling at intervals the $\theta$(horizontal) direction**

**Figure 5.** n=9 as an example, a schematic diagram of sampling points on the tangent plane

$$r(0,0) = (0,0) \tag{3}$$
$$r(\pm i, 0) = (\pm\ i \tan \Delta_\theta, 0) \tag{4}$$
$$r(0, \pm j) = (0, \pm\ j \tan \Delta_\varphi) \tag{5}$$
$$r(\pm i, \pm j) = (\pm\ i \tan\Delta_\theta, \pm\ j \tan \Delta_\varphi) \tag{6}$$

By maintaining the shape of the tangent plane at the tangent point (0,0), it is possible to displace the tangent plane along the surface of the sphere. The specific location of the tangent point $s(\theta,\varphi)$ on the sphere is arbitrarily selected, and the points on the tangent plane can be projected onto the spherical surface through an inverse transformation of the spherical center projection. This process is illustrated in Figure 6.

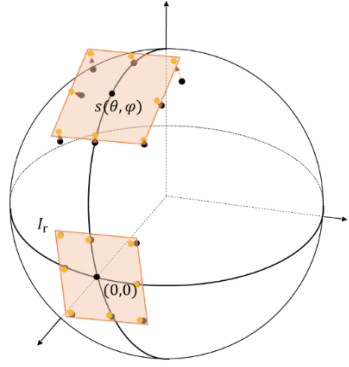

**Figure 6.** The tangent plane is displaced along the surface of the sphere

As the relative positions between the points on the tangent plane remain unchanged during the movement, their coordinates $r$ can be derived from Equ.3~Equ.6. Furthermore, the azimuth angle $\theta(i,j)$ and the zenith angle $\varphi(i,j)$ of the corresponding point mapped to the sphere can be calculated as using Equ.7~Equ.10.

$$\theta(i,j) = \theta + tan^{-1}(\frac{isinv}{\rho cos\varphi cosv - jsin\varphi sinv}) \tag{7}$$

$$\varphi(i,j) = sin^{-1}(cosvsin\varphi + \frac{jsinvcos\varphi}{\rho}) \tag{8}$$

$$\rho = \sqrt{i^2 + j^2} \tag{9}$$

$$v = tan^{-1}\rho \tag{10}$$

where $\theta$ and $\varphi$ are the coordinates of tangency.

*2.3. Transformation from perspective images to equirectangular images*

Based on the principles presented in Section 2.1 and Section 2.2, the transformation from perspective images to equirectangular images can be accomplished through the following steps:

(1) Determine the width $w$ and height $h$ of the desired equirectangular image. Calculate the sampling intervals $\Delta_\theta$ and $\Delta_\varphi$ for the horizontal and vertical directions of the equirectangular image.

(2) Adjust the size of the perspective image to $n \times n$ for convenience. Use the perspective image as the tangent plane at a point on the unit sphere $s(\theta, \varphi)$. Calculate the coordinates of each point within the $n \times n$ region of the perspective image using Equ.3~Equ.6.

(3) Employ the inverse transformation of the spherical center projection to map the sampling point $r(i,j)$ in the perspective image onto the spherical surface, obtaining the spherical coordinates of the corresponding point $s'(\theta(i,j), \varphi(i,j))$.

(4) Utilize the coordinate transformation relationship between the spherical image and the equirectangular image (Equ.1~Equ.2) to convert the spherical coordinates $s'(\theta(i,j), \varphi(i,j))$ of the mapped point into Cartesian coordinates $p(x,y)$ representing the corresponding point in the planar image. Fig.7 illustrates this process, where the yellow point $s$ represents the tangent point, the red point $r(i,j)$ represents the sampled point on the tangent plane $I_r$ image, which corresponds to the black point on the sphere $(\theta(i,j), \varphi(i,j))$, and subsequently, the points on the sphere are mapped to the red point $p(x,y)$ in the equirectangular image.

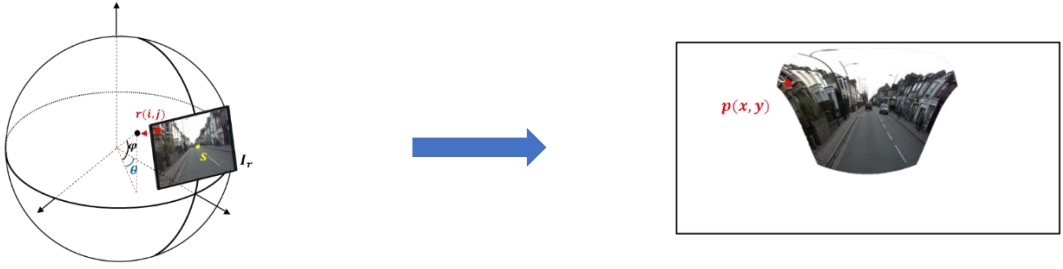

**Figure 7.** Transformation of perspective images to equirectangular images

The choice of the tangent point $s(\theta, \varphi)$ plays a significant role in determining the level of distortion in the object's shape within the equirectangular image resulting from by the transformation. When $\varphi$ approaches $\pi/2$ or $-\pi/2$ (the poles of the sphere), the object's shape becomes more distorted in the equirectangular image. Conversely, when $\varphi$ is closer to 0, the object's shape resembles its shape in the perspective image. On the other hand, the value of $\theta$ determines the horizontal position of the projection area in the equirectangular image and does not impact the object's shape. Consequently, during the conversion process, $\theta$ can be set to 0 wihout affecting the object's shape.

## 3. Datasets and Experiments

In our experiments, we utilized the CityScapes[6] and CamVid[15] datasets along with various standard CNNs semantic segmentation models. The objectives of our study are twofold:

(1) Finding the optimal projection $\varphi$ value for converting perspective images to equirectangular images: Our goal is to ensure that standard CNNs semantic segmentation models can effectively learn object distortion features by identifying the ideal projection parameter.

(2) Assessing the effectiveness of the proposed equal-angle image construction method: We aim to improve the semantic segmentation performance of standard CNNs for equirectangular images by comparing our method with alternative approaches.

Through these aims, we aim to enhance the understanding and application of semantic segmentation models for equirectangular images.

*3.1. Datasets*

CityScapes consists of street view images with a total of 34 categories, while CamVid contains street scene images with 32 categories. For our experiments, we select six common categories (roads, building, vegetation, sky, cars, and pedestrians) present in both datasets.

To transform these perspective image datasets into equirectangular images, we employ the method proposed in Section 2, resulting in the creation of the Omni-CityScapes and Omni-CamVid datasets. The transformed datasets are used for testing purposes.

We design two experimental schemes:

(1) CityScapes is used as the training set and the Omni-CamVid dataset serves as the test set.

(2) CamVid is used as the training set and the Omni-CityScapes dataset serves as the test set.

During the construction of the test set, the projection parameter φ is set to π/2, resulting in fully distorted object shapes in the test images. On the other hand, when creating the training set, $\varphi$ values within the range (0, π/2) are selected to identify the optimal $\varphi$ value that enables standard CNNs to learn effectively from the equirectangular images. The training images have a resolution of 224×224 pixels, with 700 images in the training set and 1000 images in the test set.

Our experiments focus on the upper region of the equirectangular image since the distortion degree is the same for both the upper and lower regions. Hence, we solely consider the upper region to simplify the analysis.

To ensure that the test set's image layout differs from the training set, we perform cropping on the perspective images before applying spherical projection for transformation. The cropping process is guided by the principle of projecting each category to different locations within the region where the equirectangular image is formed after spherical transformation. This ensures that the test set contains distortion at various positions.

To maintain consistency, the projected image is always positioned in the upper part of the divided equirectangular image. Consequently, any excess portion of the projected image is cropped. For instance, in Fig.8, we demonstrate this process using the car category as an example. The red box represents the area where the car is located, and it is cropped to generate a new image. Subsequently, through spherical projection, the car is positioned at different locations within the equirectangular image projection area, and any excess portion is cropped accordingly.

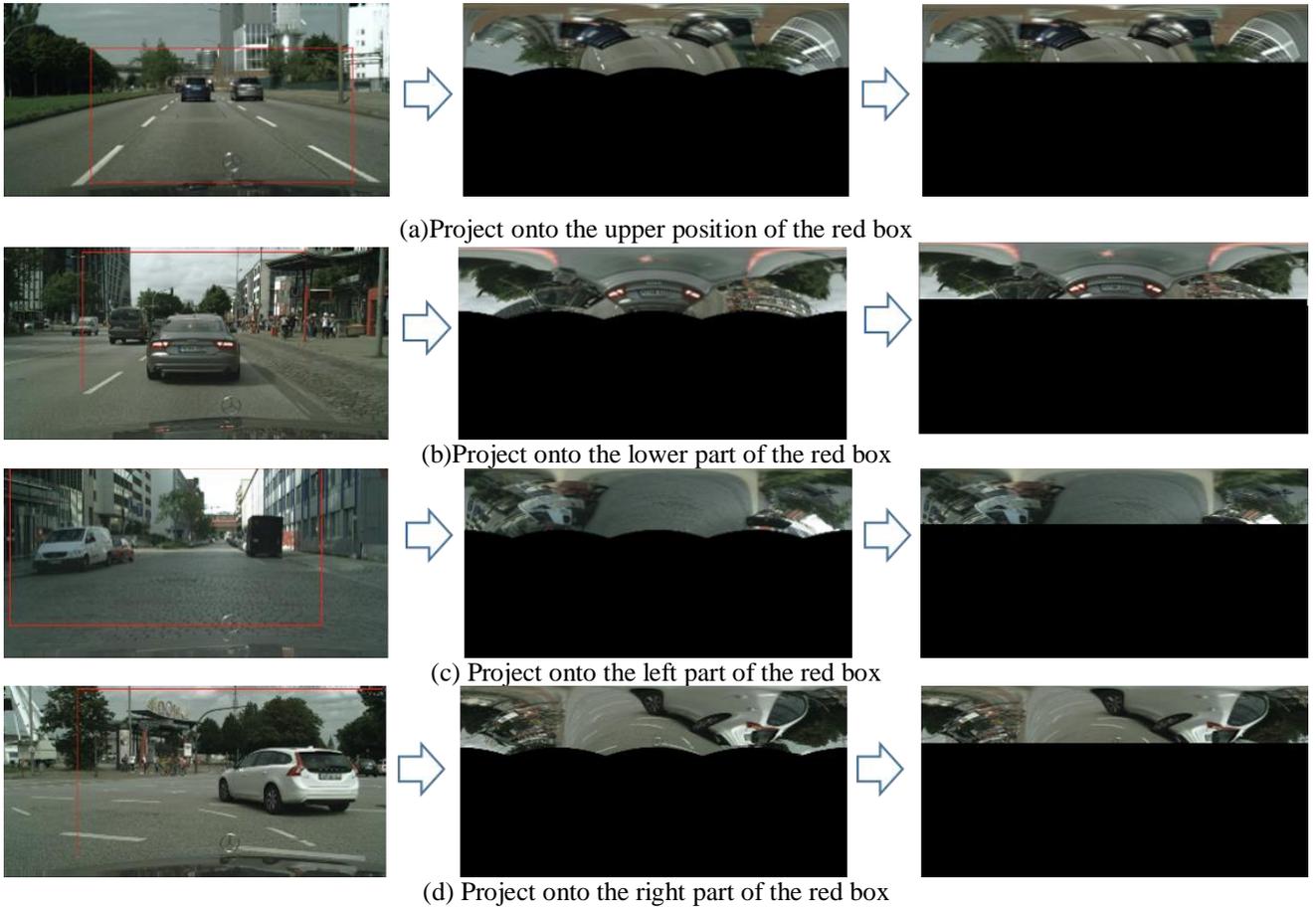

(a) Project onto the upper position of the red box

(b) Project onto the lower part of the red box

(c) Project onto the left part of the red box

(d) Project onto the right part of the red box

**Figure 8.** A transformation and cropping example of perspective images

### 3.2. Determination of the projection $\varphi$ value

(1) Experiments on different datasets with the same CNNs

For the training set, we set the projection $\varphi$ values to specific increments: $\pi/16$, $2\pi/16$, $3\pi/16$, $4\pi/16$, $5\pi/16$, $6\pi/16$, $7\pi/16$, and $8\pi/16$. The resulting equirectangular images obtained by converting the perspective images are depicted in Fig.9. It is evident that as the

$\varphi$ value increases, the degree of object shape distortion in the transformed equirectangular image becomes more pronounced. This demonstrates the impact of $\varphi$ on the level of distortion experienced by objects in the equirectangular representation.

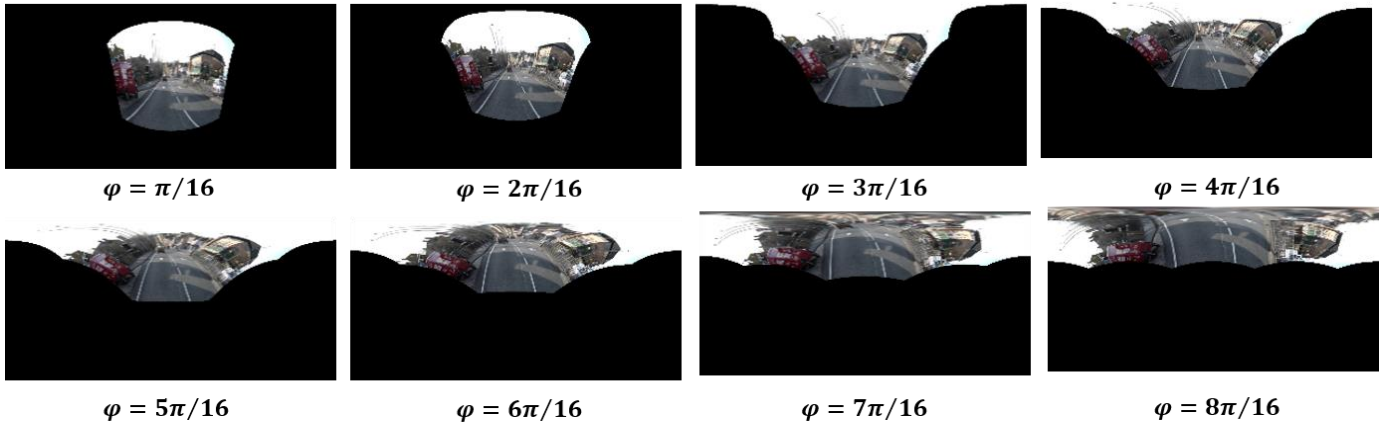

**Figure 9.** Perspective image is projected onto the plane at different locations to form equirectangular images

For the semantic segmentation model, we select the UNet[16] architecture with the VGG16 as the encoder. The VGG16 is pre-trained on the ImageNet dataset. Considering the object shape distortion in equirectangular images, we employ a specific approach in the UNet architecture. The encoder module used distorted convolutional kernels, while the decoder module utilized normal convolutional kernels. This design allowed us to analyze the impact of these different convolutional kernels on the performance of the semantic segmentation model in the context of object shape distortion.

During training, we conduct experiments for 40 epochs. The initial learning rate was set to $1 \times 10^{-4}$ for the first 25 epochs, and then adjusted to $1 \times 10^{-5}$ for the remaining epochs. The performance of the model was evaluated using the Intersection over Union (IoU) metric, a commonly used measure for semantic segmentation tasks.

The CamVid dataset is projected onto different $\varphi$ values, and the resulting equirectangular images are used to train a UNet model with distorted convolutional kernels. Table 1 presents the IoU results of this model on the Omni-CityScapes dataset. Among the different projection positions, when the perspective image is projected with a $\varphi$ value of 6π/16, the model achieves the highest average IoU value of 29.76. The IoU values for the "buildings" and "sky" categories are the highest, reaching 45.62 and 40.36, respectively. The IoU values for "roads" and "vegetation" are slightly lower than the best values, with differences of 0.26 and 0.02, respectively.

In Table 2, the results of the trained model with normal convolutional kernels are shown. Similar to the distorted convolutional kernel model, when projected with a $\varphi$ value of 6π/16, the average IoU value increases to 33.76, surpassing the other projection positions. Although the IoU values for each category are not the best, they are within a close range from the best values, differing by 0.05, 0.66, 1.26, 3.03, 0, and 7.08, respectively.

Indeed, the observation that UNet models with normal convolutional kernels outperformed those with distorted convolutional kernels is quite intriguing. This finding implies that the training dataset obtained through the transformation process can significantly enhance the model's ability to learn and understand the distorted features of objects present in equirectangular images. By utilizing normal convolutional kernels, the model appears to better capture and represent the complex spatial relationships and shape variations that arise due to the distortion inherent in equirectangular images. This highlights the importance of appropriate training data and network architecture in effectively addressing the challenges posed by object shape distortion in semantic segmentation tasks.

**Table 1.** IoU (%) of the UNet model with distorted convolution kernel when CamVid dataset is projected to different locations (best values in bold)

| $\varphi$ | roads | buildings | vegetation | sky | pedestrians | cars | average |
|---|---|---|---|---|---|---|---|
| $\pi/16$ | 27.68 | 15.63 | 9.74 | 9.13 | 0 | 6.77 | 11.49 |
| $2\pi/16$ | 33.31 | 27.31 | 19.30 | 16.03 | 0 | 4.83 | 16.80 |
| $3\pi/16$ | 36.38 | 35.68 | 17.94 | 22.13 | 0 | 7.56 | 19.95 |
| $4\pi/16$ | 37.72 | 38.43 | 27.27 | 29.63 | 0 | 10.85 | 23.98 |
| $5\pi/16$ | 37.64 | 40.00 | 44.89 | 35.09 | 0 | **12.07** | 28.28 |
| $6\pi/16$ | 38.48 | **45.62** | 47.24 | **40.36** | 0 | 6.88 | **29.76** |
| $7\pi/16$ | 38.13 | 35.93 | **47.26** | 37.47 | 0 | 6.17 | 27.49 |
| $8\pi/16$ | **38.74** | 37.19 | 45.71 | 37.50 | 0 | 3.70 | 27.14 |

**Table 2.** IoU (%) of the UNet model with normal convolution kernel when CamVid dataset is projected to different locations (best values in bold)

| $\varphi$ | roads | buildings | vegetation | sky | pedestrians | cars | Average |
|---|---|---|---|---|---|---|---|
| $\pi/16$ | 26.14 | 15.21 | 38.59 | 21.04 | 0 | 12.18 | 18.86 |
| $2\pi/16$ | 28.68 | 20.40 | 41.09 | 24.45 | 0 | 12.24 | 21.14 |
| $3\pi/16$ | 38.59 | 47.88 | 42.14 | 28.80 | 0 | **19.44** | 29.48 |
| $4\pi/16$ | 39.03 | 46.40 | 52.81 | 35.72 | 0 | 11.05 | 30.84 |
| $5\pi/16$ | **39.41** | **52.16** | 54.47 | 39.06 | 0 | 11.13 | 32.71 |
| $6\pi/16$ | 39.36 | 51.50 | 58.28 | 41.08 | 0 | 12.36 | **33.76** |
| $7\pi/16$ | 38.96 | 43.02 | **59.54** | 37.77 | 0 | 13.65 | 32.16 |
| $8\pi/16$ | 38.94 | 47.07 | 58.00 | **44.11** | 0 | 12.08 | 33.37 |

The results presented in Table 3 demonstrate the performance of the UNet model with distorted convolutional kernels trained on CityScapes dataset, projected to various locations, and tested on Omni-CamVid. Notably, when the perspective image is projected with a $\varphi$ value of $6\pi/16$, the model achieves the highest average IoU value of 23.96. Furthermore, the IoU values for roads, buildings, and sky are also higher compared to other projection positions, reaching 31.87, 28.02, and 54.90, respectively. However, the IoU values for vegetation, pedestrians, and cars fall short of the highest values by 1.29, 0, and 1.84, respectively.

On the other hand, Table 4 presents the segmentation results of the UNet model with normal convolutional kernels trained on CityScapes dataset and tested on Omni-CamVid, with images projected to different locations. Notably, when projected to $4\pi/16$, buildings achieve the highest IoU value of 30.92, while cars achieve the highest IoU value of 7.64. When projected to $8\pi/16$, roads and vegetation achieve the highest IoU values of 30.96 and 29.15, respectively. Additionally, when projected to $6\pi/16$, the sky exhibits the highest IoU value of 58.89. In terms of the average IoU value for all six categories, projecting to $6\pi/16$ yields the highest value of 25.03, surpassing other projection positions. However, there are slight differences between the highest IoU values and the IoU values obtained for roads, buildings, vegetation, pedestrians, and cars, with variations of 0.87, 3.50, 1.29, 0, and 1.70, respectively.

Comparing the results presented in Table 3 and Table 4, it is evident that, across the eight different projection positions, the average IoU of the model utilizing normal convolutional kernels consistently outperforms that of the model employing distorted convolutional kernels. This finding aligns with the observation from previous experiments, reinforcing the notion that employing common convolutional kernels facilitates improved performance in handling the challenges associated with object shape distortion in semantic segmentation tasks.

Table 3. IoU (%) of the UNet model with distorted convolution kernel when CityScapes dataset is projected to different locations (best values in bold)

| $\varphi$ | roads | buildings | vegetation | sky | pedestrians | cars | average |
|---|---|---|---|---|---|---|---|
| $\pi/16$ | 24.26 | 1.68 | 0.84 | 8.23 | 0 | 4.23 | 6.54 |
| $2\pi/16$ | 25.13 | 7.35 | 1.87 | 17.95 | 0 | 4.10 | 9.40 |
| $3\pi/16$ | 28.02 | 13.52 | 11.32 | 16.26 | 0 | 6.02 | 12.52 |
| $4\pi/16$ | 31.07 | 22.26 | 11.39 | 49.28 | 0 | **7.69** | 20.28 |
| $5\pi/16$ | 31.29 | 26.38 | **24.42** | 46.96 | 0 | 6.61 | 22.61 |
| $6\pi/16$ | **31.87** | **28.02** | 23.13 | **54.90** | 0 | 5.85 | **23.96** |
| $7\pi/16$ | 29.07 | 16.04 | 24.12 | 45.39 | 0 | 6.36 | 20.16 |
| $8\pi/16$ | 28.76 | 16.76 | 23.12 | 27.33 | 0 | 7.12 | 17.18 |

Table 4. IoU (%) of the UNet model with normal convolution kernel when CityScapes dataset is projected to different locations (best values in bold)

| $\varphi$ | roads | buildings | vegetation | sky | pedestrians | cars | average |
|---|---|---|---|---|---|---|---|
| $\pi/16$ | 23.27 | 4.14 | 2.20 | 19.39 | 0 | 4.88 | 8.98 |
| $2\pi/16$ | 24.61 | 4.79 | 2.91 | 26.15 | 0 | 5.36 | 10.64 |
| $3\pi/16$ | 25.79 | 30.61 | 18.21 | 51.55 | 0 | 5.24 | 21.90 |
| $4\pi/16$ | 27.19 | **30.92** | 16.98 | 47.12 | 0 | **7.64** | 21.64 |
| $5\pi/16$ | 28.26 | 27.93 | 24.34 | 55.89 | 0 | 6.32 | 23.79 |
| $6\pi/16$ | 30.09 | 27.42 | 27.86 | **58.89** | 0 | 5.94 | **25.03** |
| $7\pi/16$ | 29.48 | 22.16 | 25.01 | 41.45 | 0 | 6.19 | 20.72 |
| $8\pi/16$ | **30.96** | 30.31 | **29.15** | 38.26 | 0 | 5.13 | 22.30 |

Based on the analysis of the IoU results presented in Table 1-4, it is observed that the highest average IoU value is achieved when the perspective images are projected to approximately $6\pi/16$. Subdividing the φ value further does not significantly impact the IoU values, as the differences become very small. Hence, $6\pi/16$ can be considered as the approximate optimal projection position.

To provide visual evidence of the segmentation results, Fig. 10 and Fig. 11 illustrate the segmentation outputs of the UNet model with normal convolutional kernels trained on the CamVid and CityScapes datasets, respectively, using different projection $\varphi$ values. The visualization results indicate that when the $\varphi$ value is set to $6\pi/16$, the overall segmentation performance of the model is superior compared to other projection positions. However, it is worth noting that the segmentation performance for small-sized objects, such as pedestrians and cars, remains challenging in perspective image processing tasks. This finding highlights an area that requires further improvement and attention in future research.

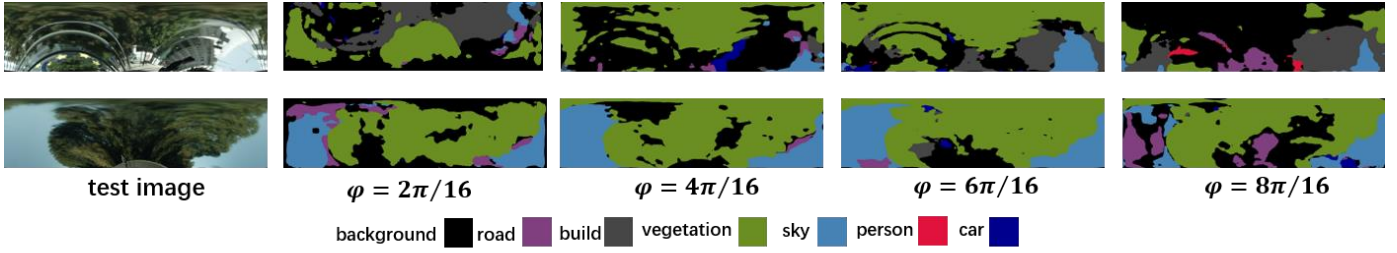

**Figure 10.** The segmentation results of UNet model with normal convolution kernel trained on CamVid dataset

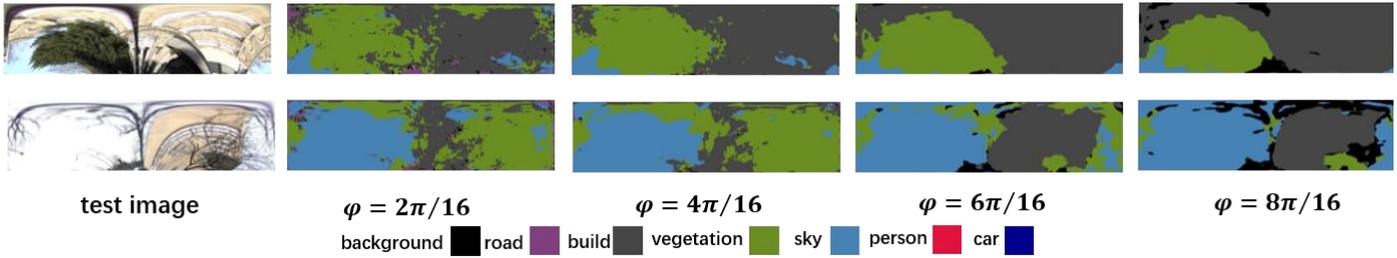

**Figure 11.** The segmentation results of UNet model with normal convolution kernel trained on CityScapes dataset

(2) Same dataset, different CNNs

The experimental setup involves using the CamVid dataset as the training set and the Omni-CityScapes dataset as the testing set. Various CNN models, including UNet, UNet++[17], SegNet[18], PSPNet[19], and DeepLab v3+[20], are utilized with normal convolutional kernels. The results of these experiments are presented in Tables 5 to 9. Based on the segmentation results shown in Fig.13 to Fig.17, it can be observed that when the $\varphi$ value is set to $6\pi/16$, the average IoU values obtained using UNet, SegNet, and PSPNet models are higher compared to other $\varphi$ values. However, for UNet++ and DeepLab v3+ models, the highest average IoU value is achieved at a $\varphi$ value of $5\pi/16$. Nevertheless, the difference in IoU values between $5\pi/16$ and $6\pi/16$ is very small, with only 0.03 and 0.33 differences, respectively. This suggests that even though some models do not achieve the highest IoU at the $6\pi/16$ projection position, the difference is minimal. Hence, for different semantic segmentation models, the approximation of $6\pi/16$ can be considered as the actual optimal projection position.

**Table 5.** IoU(%) of UNet network (backbone network is VGG16 and ResNet34)

| $\varphi$ | backbone | roads | buildings | vegetation | sky | pedestrians | cars | average |
|---|---|---|---|---|---|---|---|---|
| $\pi/16$ | VGG16 | 26.14 | 15.21 | 38.59 | 21.04 | 0 | 12.18 | 18.86 |
|  | ResNet34 | 9.76 | 7.60 | 3.51 | 17.09 | 0 | 3.19 | 6.86 |
| $2\pi/16$ | VGG16 | 28.68 | 20.40 | 41.09 | 24.45 | 0 | 12.24 | 21.14 |
|  | ResNet34 | 12.15 | 11.77 | 4.64 | 30.98 | 0 | 1.59 | 10.19 |
| $3\pi/16$ | VGG16 | 38.59 | 47.88 | 42.14 | 28.80 | 0 | **19.44** | 29.48 |
|  | ResNet34 | 31.30 | 23.80 | 33.03 | 22.97 | 0 | 7.95 | 19.84 |
| $4\pi/16$ | VGG16 | 39.03 | 46.40 | 52.81 | 35.72 | 0 | 11.05 | 30.84 |
|  | ResNet34 | 35.34 | 38.75 | 39.24 | 28.27 | 0 | 4.74 | 24.39 |
| $5\pi/16$ | VGG16 | **39.41** | **52.16** | 54.47 | 39.06 | 0 | 11.13 | 32.71 |
|  | ResNet34 | 37.89 | 43.01 | 31.90 | 36.16 | 0 | 4.82 | 25.63 |

| $\varphi$ | backbone | roads | buildings | vegetation | sky | pedestrians | cars | average |
|---|---|---|---|---|---|---|---|---|
| $\pi/16$ | VGG16 | 26.14 | 15.21 | 38.59 | 21.04 | 0 | 12.18 | 18.86 |
|  | ResNet34 | 9.76 | 7.60 | 3.51 | 17.09 | 0 | 3.19 | 6.86 |
| $2\pi/16$ | VGG16 | 28.68 | 20.40 | 41.09 | 24.45 | 0 | 12.24 | 21.14 |
|  | ResNet34 | 12.15 | 11.77 | 4.64 | 30.98 | 0 | 1.59 | 10.19 |
| $3\pi/16$ | VGG16 | 38.59 | 47.88 | 42.14 | 28.80 | 0 | **19.44** | 29.48 |
|  | ResNet34 | 31.30 | 23.80 | 33.03 | 22.97 | 0 | 7.95 | 19.84 |
| $6\pi/16$ | VGG16 | 39.36 | 51.50 | 58.28 | 41.08 | 0 | 12.36 | **33.76** |
|  | ResNet34 | 38.66 | 42.70 | 42.76 | 35.15 | 0 | 3.68 | 27.16 |
| $7\pi/16$ | VGG16 | 38.96 | 43.02 | **59.54** | 37.77 | 0 | 13.65 | 32.16 |
|  | ResNet34 | 39.07 | 45.83 | 44.69 | 39.38 | 0 | 7.59 | 29.43 |
| $8\pi/16$ | VGG16 | 38.94 | 47.07 | 58.00 | **44.11** | 0 | 12.08 | 33.37 |
|  | ResNet34 | 37.74 | 31.02 | 44.17 | 31.88 | 0 | 8.17 | 25.50 |

**Table 6.** IoU(%) of UNet++ network (backbone network is VGG16 and ResNet34)

| $\varphi$ | backbone | roads | buildings | vegetation | sky | pedestrians | cars | average |
|---|---|---|---|---|---|---|---|---|
| $\pi/16$ | VGG16 | 24.37 | 17.04 | 15.55 | 25.11 | 0 | 6.16 | 14.71 |
|  | ResNet34 | 19.31 | 18.41 | 19.15 | 22.80 | 0 | 7.89 | 14.59 |
| $2\pi/16$ | VGG16 | 22.23 | 19.66 | 28.84 | 32.73 | 0 | 5.57 | 18.17 |
|  | ResNet34 | 19.44 | 6.04 | 3.58 | 27.08 | 0 | 5.89 | 10.34 |
| $3\pi/16$ | VGG16 | 35.93 | 47.66 | 39.40 | 36.30 | 0 | 11.65 | 28.49 |
|  | ResNet34 | 32.07 | 34.41 | 36.73 | 27.20 | 0 | 9.43 | 23.31 |
| $4\pi/16$ | VGG16 | 37.85 | **49.58** | 53.40 | 32.47 | 0 | 7.35 | 30.11 |
|  | ResNet34 | 33.55 | 35.05 | 38.17 | 37.70 | 0 | 8.70 | 25.53 |
| $5\pi/16$ | VGG16 | **39.08** | 48.80 | **55.57** | 34.75 | 0 | 6.70 | **30.82** |
|  | ResNet34 | 38.41 | 46.30 | 40.45 | 38.67 | 0 | 9.07 | 28.82 |
| $6\pi/16$ | VGG16 | 38.66 | 46.99 | 52.06 | 35.03 | 0 | 12.01 | 30.79 |
|  | ResNet34 | 38.74 | 36.40 | 48.42 | 35.07 | 0 | **15.08** | 28.95 |
| $7\pi/16$ | VGG16 | 37.71 | 37.71 | 52.29 | 32.12 | 0 | 11.43 | 28.54 |
|  | ResNet34 | 38.65 | 33.71 | 47.87 | 34.52 | 0 | 10.68 | 27.57 |
| $8\pi/16$ | VGG16 | 37.80 | 37.47 | 52.31 | **38.78** | 0 | 9.71 | 29.35 |
|  | ResNet34 | 38.58 | 29.41 | 50.67 | 36.03 | 0 | 13.52 | 28.04 |

**Table 7.** IoU(%) of SegNet network (backbone network is VGG16)

| $\varphi$ | roads | buildings | vegetation | sky | pedestrians | cars | average |
|---|---|---|---|---|---|---|---|
| $\pi/16$ | 13.31 | 1.74 | 0 | 0 | 0 | 0 | 2.51 |
| $2\pi/16$ | 20.42 | 15.26 | 0.38 | 0.33 | 0 | 2.12 | 6.42 |
| $3\pi/16$ | 27.54 | 27.34 | 2.00 | 1.02 | 0 | 0.75 | 9.78 |
| $4\pi/16$ | 37.52 | 38.15 | 21.81 | 3.15 | 0 | 1.06 | 16.95 |
| $5\pi/16$ | 37.90 | 33.56 | 4.13 | 13.18 | 0 | 3.65 | 15.40 |
| $6\pi/16$ | **38.46** | **40.69** | **37.55** | 22.94 | 0 | **8.44** | **24.68** |
| $7\pi/16$ | 37.06 | 25.80 | 6.63 | 21.23 | 0 | 6.28 | 16.17 |

| $\varphi$ | roads | buildings | vegetation | sky | pedestrians | cars | average |
|---|---|---|---|---|---|---|---|
| $8\pi/16$ | 37.48 | 17.62 | 24.61 | **30.13** | 0 | 3.93 | 18.96 |

Table 8. IoU(%) of PSPNet network (backbone network is VGG16 and ResNet34)

| $\varphi$ | backbone | roads | buildings | vegetation | sky | pedestrians | cars | average |
|---|---|---|---|---|---|---|---|---|
| $\pi/16$ | VGG16 | 5.56 | 0.15 | 0.20 | 0.35 | 0 | 0 | 1.04 |
| | ResNet34 | 8.28 | 14.68 | 2.30 | 1.76 | 0 | 0.04 | 4.51 |
| $2\pi/16$ | VGG16 | 9.07 | 5.29 | 11.74 | 3.20 | 0 | 0 | 4.88 |
| | ResNet34 | 18.99 | 34.52 | 23.03 | 9.87 | 0 | 0.18 | 14.43 |
| $3\pi/16$ | VGG16 | 29.68 | 48.73 | 29.61 | 8.51 | 0 | 0 | 19.42 |
| | ResNet34 | 35.71 | 47.04 | 9.80 | 23.02 | 0 | 0 | 19.26 |
| $4\pi/16$ | VGG16 | 33.97 | **52.35** | 43.24 | 13.66 | 0 | 0 | 23.87 |
| | ResNet34 | 35.46 | 50.54 | 31.83 | 23.90 | 0 | 0.26 | 23.67 |
| $5\pi/16$ | VGG16 | 30.30 | 48.70 | 46.42 | 19.66 | 0 | 1.01 | 24.35 |
| | ResNet34 | 36.96 | 52.20 | 47.73 | 32.88 | 0 | 0.46 | 28.37 |
| $6\pi/16$ | VGG16 | 33.22 | 52.14 | 37.02 | 29.10 | 0 | 1.09 | 25.43 |
| | ResNet34 | 37.82 | 51.09 | **55.81** | 32.90 | 0 | **3.74** | **30.23** |
| $7\pi/16$ | VGG16 | 35.29 | 50.59 | 53.83 | 33.00 | 0 | 3.66 | 29.40 |
| | ResNet34 | 37.76 | 42.36 | 53.24 | 29.44 | 0 | 3.38 | 27.70 |
| $8\pi/16$ | VGG16 | 35.77 | 44.72 | 48.20 | 34.38 | 0 | 1.06 | 27.36 |
| | ResNet34 | **38.12** | 42.37 | 53.67 | **36.78** | 0 | 2.84 | 28.96 |

Table 9. IoU(%) of DeepLab v3+ network (backbone network is ResNet34)

| $\varphi$ | roads | buildings | vegetation | sky | pedestrians | cars | average |
|---|---|---|---|---|---|---|---|
| $\pi/16$ | 7.91 | 4.72 | 1.01 | 19.67 | 0 | 0.3 | 5.60 |
| $2\pi/16$ | 27.26 | 30.85 | 27.10 | 26.75 | 0 | 3.14 | 19.18 |
| $3\pi/16$ | 37.21 | 29.72 | 28.50 | 24.92 | 0 | 7.20 | 21.26 |
| $4\pi/16$ | 35.92 | **49.62** | 39.01 | 34.52 | 0 | 7.62 | 27.78 |
| $5\pi/16$ | 37.76 | 49.05 | 36.87 | **36.44** | 0 | 10.87 | **28.50** |
| $6\pi/16$ | **38.65** | 43.36 | 48.42 | 31.75 | 0 | 6.84 | 28.17 |
| $7\pi/16$ | 37.93 | 35.12 | **50.47** | 31.36 | 0 | 6.41 | 26.88 |
| $8\pi/16$ | 37.11 | 18.48 | 46.33 | 35.98 | 0 | **11.15** | 24.84 |

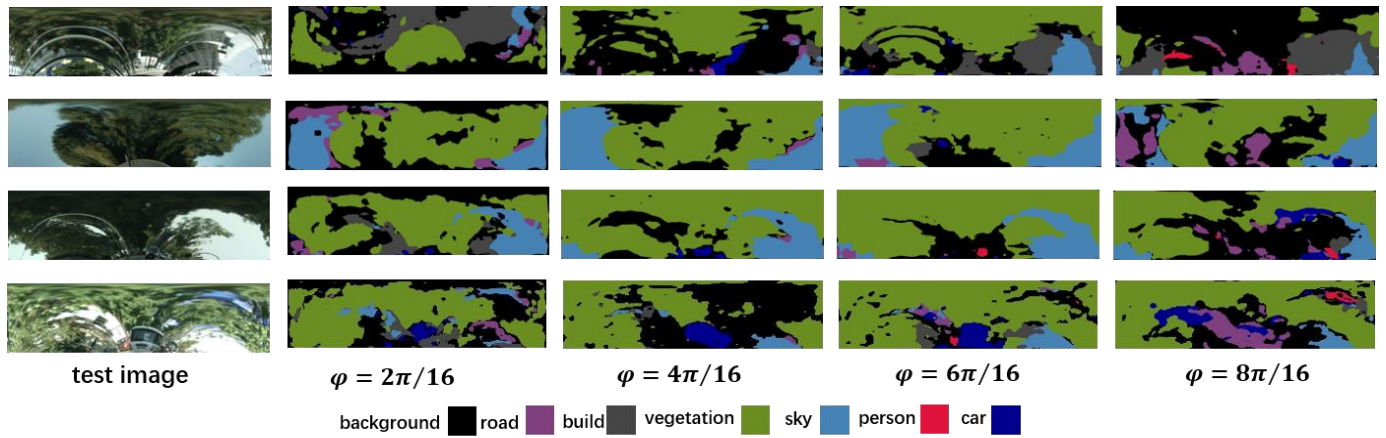

**Figure 12.** The segmentation results of UNet with backbone VGG 16

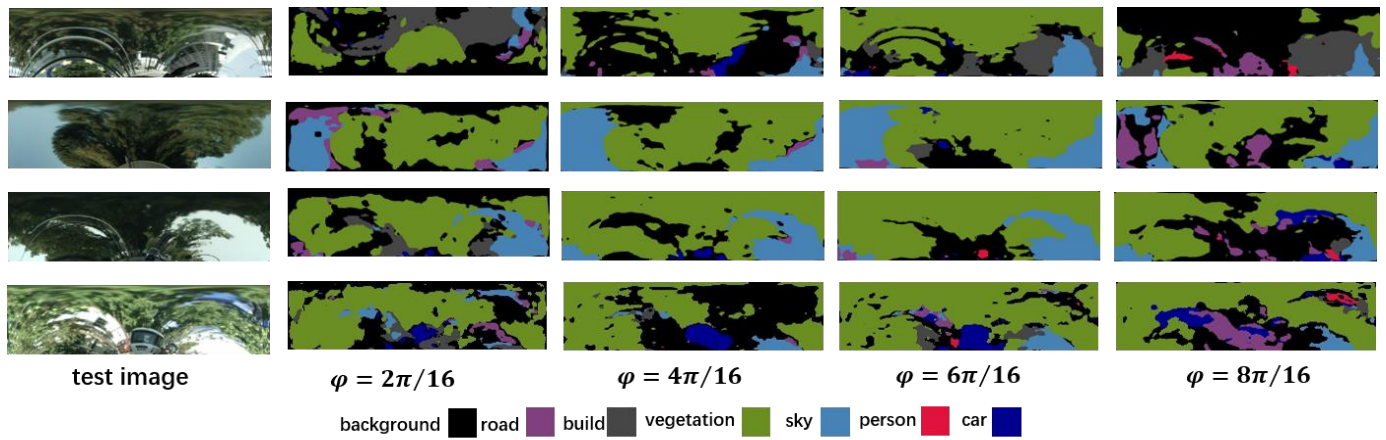

**Figure 13.** The segmentation results of UNet++ with backbone VGG 16

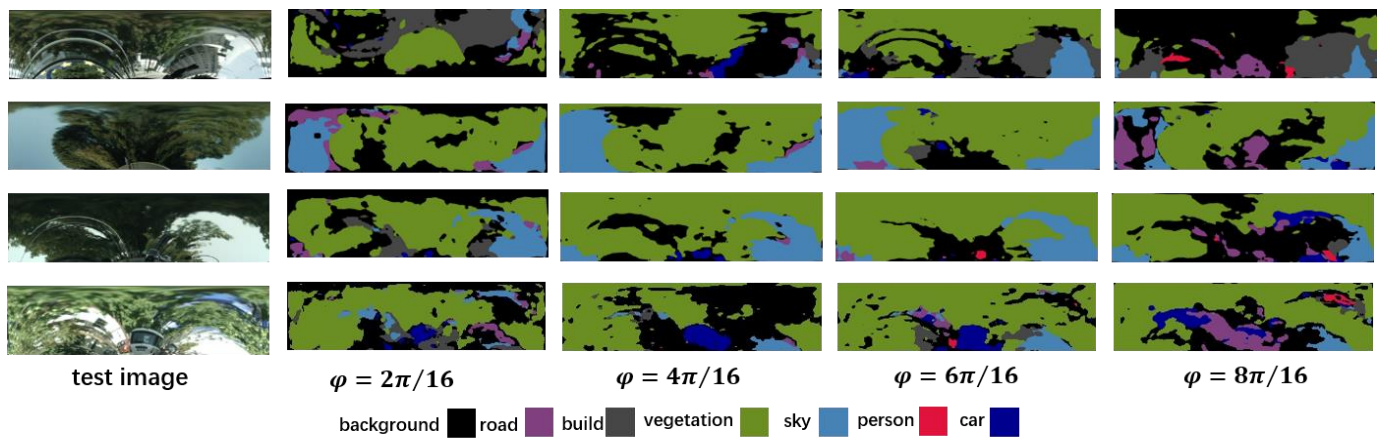

**Figure 14.** The segmentation results of SegNet with backbone VGG 16

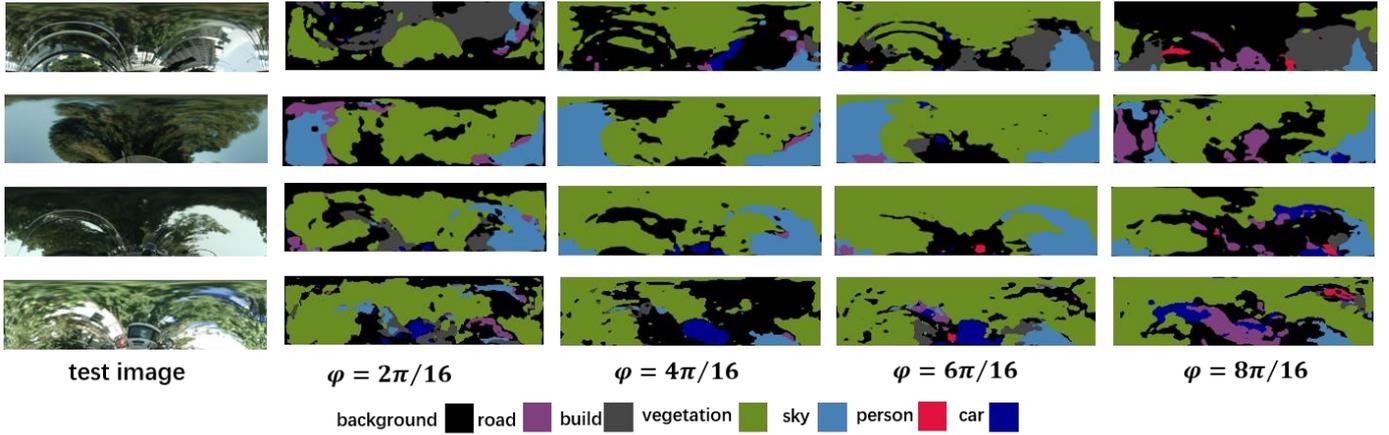

**Figure 15.** The segmentation results of PSPNet with backbone ResNet34

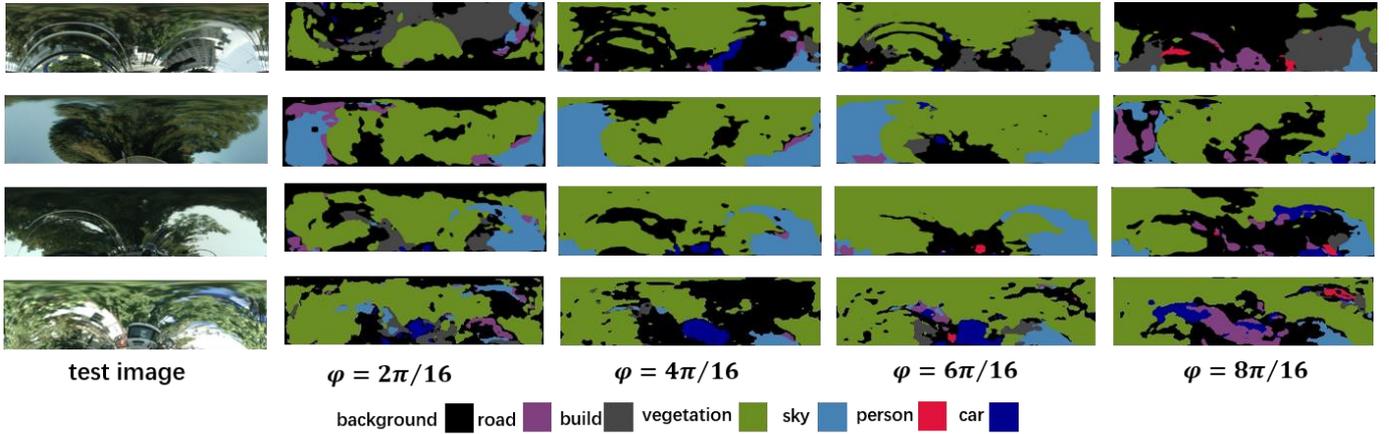

**Figure 16.** The segmentation results of DeepLab v3+ with backbone ResNet34

*3.3 Methods Comparison*

In the comparison experiments, three different approaches, namely supervised learning, unsupervised learning, and data augmentation, are evaluated alongside the method proposed in the paper. The experiments are conducted using the CamVid dataset as the training set and the Omni-Cityscapes dataset as the test set, with 700 and 1000 images, respectively. The UNet model is chosen as the semantic segmentation model, with VGG16 as the backbone network.

(1) Tangent plane image method [10]: This approach involves using normal convolutional kernels for training and distorted convolutional kernels for testing. Tangent plane images are generated from equirectangular images in the Omni-Cityscapes dataset using spherical center projection. These tangent plane images are then used as input for training the UNet network. During testing, the weights of the normal convolutional kernels in the UNet network are copied to the distorted convolutional kernels for segmenting equirectangular images. This method utilizes the labeled Omni-Cityscapes dataset, making it a form of supervised learning.

(2) UNet-P2PDA [12]: This method employs an adversarial generation network that takes both perspective images and equirectangular images as input. The goal is to minimize the difference between the segmentation results of the two types of images. Since no labeled equirectangular images are required for training, this approach falls under the category of unsupervised learning.

(3) Image Enhancement: This method involves performing operations such as cropping, upscaling, and left/right rotations on the perspective images to increase the size of the dataset and improve the learning capability of the UNet model.

The method proposed in the paper uses normal convolutional kernels for both training and testing. The experimental results of this method, as well as the three comparison methods mentioned above, are presented in Table 10.

Table 10. Comparison of the method in this paper with other methods (IoU)

| Methods | roads | buildings | vegetation | sky | pedestrians | cars | Average |
|---|---|---|---|---|---|---|---|
| Tangent plane image method | 28.07 | 21.31 | 32.37 | 34.55 | 0.12 | 3.03 | 19.91 |
| UNet-P2PDA | 30.80 | 38.58 | 34.87 | 27.38 | 0 | 6.74 | 23.06 |
| Image enhancement | **42.80** | 35.85 | 35.52 | **42.04** | 0.18 | 2.76 | 26.53 |
| Method in this paper($\varphi = 6\pi/16$) | 39.36 | **51.50** | **58.28** | 41.08 | 0 | **12.36** | **33.76** |

Table 10 demonstrates that the method proposed in the paper achieves the highest IoU values compared to the other three methods, with an average IoU of 33.76. Therefore, the method in this paper outperforms the other methods in terms of overall segmentation effectiveness. Specifically, the tangent plane image method has a significantly lower average IoU value of 19.91. This can be attributed to the generation of numerous similar tangent plane images, leading to overfitting of the model and reduced generalization ability. Consequently, the model performs poorly on the test dataset, resulting in lower IoU values and inferior segmentation performance. For the UNet-P2PDA method, the average IoU value is 23.06, indicating that the approach of using an adversarial generation network is less effective in handling the upper and lower regions of equirectangular images. Regarding the image enhancement method, the model achieves the highest IoU values of 42.8 and 42.04 for roads and sky, respectively. It is noteworthy that the training set processing procedure of this method can be seen as a specialized form of image enhancement, specifically targeted at learning distortion features of the upper and lower regions of equirectangular images. Consequently, the model can better capture the distorted features of these regions and achieve superior performance. When considering individual categories, the method in this paper achieves the highest IoU values of 51.50, 58.28, and 12.36 for buildings, vegetation, and cars, respectively. This is attributed to the projection of perspective images onto equirectangular images, enabling the model to learn the distorted features of object shapes during training and improving its semantic segmentation capability.

## 4. Conclusion

360-degree spherical images offer the advantage of a wide field of view and are commonly projected onto a planar plane as an equirectangular image for further processing. However, the object shapes in equirectangular images can be distorted, and they often lack translation invariance. Moreover, there is a scarcity of publicly available datasets containing labeled equirectangular images, which poses a challenge for standard convolutional neural network (CNN) models to effectively process such images.

To address these challenges, we propose a methodology that involves converting a perspective image into equirectangular image. This is achieved through the use of inverse transformations, specifically the spherical center projection and the equidistant cylindrical projection. By employing these transformations, standard CNN models can learn the distortion features at different positions within the equirectangular image, enabling them to perform semantic segmentation effectively.

The parameter $\varphi$, which determines the projection position of the perspective image, has been thoroughly analyzed using various datasets and CNN models such as UNet,

UNet++, SegNet, PSPNet, and DeepLab v3+. The experimental results consistently demonstrate that an optimal value of φ for achieving effective semantic segmentation of equirectangular images using standard CNNs is 6π/16.

Comparing our proposed method with three other types of approaches, namely supervised learning, unsupervised learning, and data augmentation, we find that our method outperforms them in terms of average Intersection over Union (IoU) value. Specifically, our method achieves an average IoU value of 43.76%, which is significantly higher than the corresponding values of the other three methods. It surpasses them by 23.85%, 10.7%, and 17.23% respectively, highlighting the superiority of our proposed methodology in semantically segmenting equirectangular images.

In future research, we intend to explore the adaptability of our proposed methodology to various other tasks such as isometric image classification, object detection, and depth prediction. By extending our methodology to these areas, we hope to contribute to the broader field of computer vision and enable more comprehensive analysis and understanding of 360-degree spherical images.

**Author Contributions:** The authors Haoqian Chen, Rencheng Sun and Yi Sui developed the initial idea for the study. The authors Haoqian Chen, Jian Liu, Minghe Li designed the research methodology, including data collection, experimental procedures. The authors Kaiwen Jiang, Minghe Li and Ziheng Xu created figures, tables of the presentation of results. The authors Haoqianchen and Yi Sui wrote the initial version of the manuscript. The authors Kaiwen Jiang and Yi Sui revised and edited the manscript. The author Yi Sui provided oversight and guidance througout the research. All authors commented on previous versions of the manuscript. All authors read and approved the final manuscript.

**Funding:** This research was supported by Young Scientists Fund of the National Natural Science Foundation of China (41706198) and Qingdao Independent innovation major special project (21-1-2-1hy).

**Data Availability Statement:** All data generated or analyzed during this study are included in this published article.

**Conflicts of Interest:** All authors certify that they have no affiliations with or involvement in any organization or entity with any financial interest or non-financial interest in the subject matter or materials discussed in this manuscript.